%% file: EntropyInfer.tex
\documentclass[11pt,fleqn]{article}

\usepackage[preprint]{acl}

\usepackage{times}
\usepackage{latexsym}
\usepackage{multirow}
\usepackage{makecell}

 \usepackage[normalem]{ulem}
 \useunder{\uline}{\ul}{}

\usepackage[T1]{fontenc}

\usepackage[utf8]{inputenc}

\usepackage{microtype}

\usepackage{inconsolata}

\usepackage{graphicx}

\usepackage{xcolor}
\usepackage{algorithm}
\usepackage{algpseudocode}
\usepackage{amsmath}
\usepackage{amssymb}
\usepackage{subcaption}
\usepackage{enumitem}
\usepackage{hyperref}
\providecommand{\tightlist}{%
	\setlength{\itemsep}{0pt}\setlength{\parskip}{0pt}}
\title{From Rigid to Dynamic: Entropy-Guided Adaptive Inference for Long-Context LLMs}

\author{
  \textbf{Zhanchao Xu}\textsuperscript{1},
  \textbf{Haoyang Li}\textsuperscript{1},
  \textbf{Qingfa Xiao}\textsuperscript{2},
  \textbf{Fei Teng}\textsuperscript{3},
  \\
  \textbf{Chen Jason Zhang}\textsuperscript{1},
  \textbf{Lei Chen}\textsuperscript{2},
  \textbf{Qing Li}\textsuperscript{1}
  \\
  \textsuperscript{1}Department of Computing, PolyU \quad
  \textsuperscript{2}DSA, HKUST(GZ) \quad
  \textsuperscript{3}CSE, HKUST \\
  \small{\texttt{zhanchaoxu0228@gmail.com}, \texttt{\{haoy1li, csqli\}@comp.polyu.edu.hk}, \texttt{jason-c.zhang@polyu.edu.hk}} \\
  \small{\texttt{qxiao183@connect.hkustgz.edu.cn}, \texttt{fteng@connect.ust.hk}, \texttt{leichen@cse.ust.hk}}
}

\begin{document}
\maketitle
\begin{abstract}
Existing sparse attention and KV cache compression methods for long-context LLM inference typically apply fixed sparsity patterns or uniform budgets across all attention heads, overlooking the substantial variation in attention behavior among heads and contexts. We observe two distinct entropy patterns among attention heads: \emph{Rigid Heads}, whose entropy stays near zero across input segments, and \emph{Dynamic Heads}, whose entropy fluctuates significantly. Crucially, the distribution of these types is context-dependent and cannot be predetermined offline. We therefore propose EntropyInfer, a training-free framework that uses attention entropy to adaptively allocate compute at the granularity of individual heads and segments during prefilling. For decoding, we introduce a latent KV cache compression scheme that leverages generated output tokens, rather than prefill tokens alone, to identify and retain the most critical cache entries. Extensive experiments on Llama, Qwen and openPangu model series show that EntropyInfer consistently outperforms baselines including SnapKV, AdaKV, and CritiPrefill, achieving up to 2.39$\times$
end-to-end speedup beyond 100k tokens with minimal quality degradation compared to full attention.
The code is released in \url{https://github.com/SHA-4096/EntropyInfer}.
\end{abstract}

\input{section/sec-introduction.tex}
\input{section/sec-preliminary-and-related-work.tex}

\input{section/sec-motivation.tex}
\input{section/sec-method.tex}

\input{section/sec-experiment.tex}

\input{section/sec-conclusion.tex}
 
\input{section/sec-limitation.tex}

\bibliography{custom}

\clearpage
\onecolumn

\input{section/sec-appendix.tex}

\end{document}

%% file: section/sec-introduction.tex
\section{Introduction}\label{sec:introduction}

Long-context inference has emerged as a critical bottleneck for modern LLM deployment. As context lengths grow into the long-context regime, two cost centers dominate end-to-end latency: the quadratic attention computation during prefilling, and the linearly growing KV cache during decoding. Reducing either cost typically requires discarding information from the attention computation or the cache, creating a persistent tension between inference efficiency and generation quality. We show that attention entropy, measured per head and per segment during inference, exposes structure that existing methods fail to exploit and enables inference acceleration with negligible quality loss.

Attention computation overhead during prefilling grows quadratically with context length and dominates first-token latency in long-context scenarios. Existing approaches accelerate prefilling by computing only a subset of the attention matrix, and the strategies for choosing that subset fall into three categories, each with a structural limitation. Methods with predefined sparsity patterns (e.g., MInference~\citep{jiangMInference10Accelerating2024}) assume a fixed geometric structure that cannot adapt to context-dependent attention shifts. Globally adaptive methods (e.g., FlexPrefill~\citep{laiFlexPrefillCONTEXTAWARESPARSE2025}) tune a single sparsity threshold across all heads, ignoring head-level heterogeneity. Block-level criticality methods (e.g., CritiPrefill~\citep{lvCritiPrefillSegmentwiseCriticalitybased2024}) allocate a uniform per-segment budget, treating every head as equally informative. All three families assume a form of homogeneity, across positions, across heads, or across contexts, that we find empirically does not hold.

During decoding, the linearly growing KV cache becomes the dominant memory pressure on the GPU. Eviction-based methods such as SnapKV~\citep{liSnapKVLLMKnows2024} and AdaKV~\citep{fengAdaKVOptimizingKV2024} compress the cache by retaining only tokens deemed important under the prefill-stage attention pattern. This design has a fundamental blind spot: recent studies~\citep{liLoopServeAdaptiveDualphase2025, wuLouisKVEfficientKV2025} show that the attention pattern shifts substantially once generation begins, so tokens that look important under the prefill signal are not reliably the tokens that drive the decoding process. Permanently evicting tokens at the boundary between the prefill and decode stage therefore risks discarding precisely the entries the generator will later need.

A complementary line of work recognizes that different attention heads play different roles, as observed in RazorAttention~\citep{tangRazorAttentionEfficientKV2024}. However, RazorAttention and similar head-aware methods rely on offline profiling over a calibration set, baking head categories into static configurations. Through a per-head entropy analysis on Llama-3.1-8B-Instruct and Qwen-2.5-7B-Instruct, we find two phenomena that make offline categorization fundamentally limited. First, attention heads cleanly separate into two regimes: \emph{Rigid Heads}, whose row-wise entropy stays below $10^{-5}$ regardless of input and whose attention is therefore near-deterministic, and \emph{Dynamic Heads}, whose entropy fluctuates substantially across query positions and carries genuine context-dependent semantic structure. Second, the assignment of a given head to one regime is itself context-dependent, meaning the same head can behave as Rigid on one input and Dynamic on another. Together, these observations imply that head categorization must happen online, and that entropy fluctuation itself provides a cheap online signal for allocating compute.

Building on this insight, we propose \textbf{EntropyInfer}, a training-free framework that uses attention entropy as an online signal to allocate inference cost adaptively across both stages. During prefilling, we estimate per-head, per-segment entropy from a low-cost observation attention matrix; Rigid heads receive a fixed minimal budget, while Dynamic heads receive a budget that scales with entropy fluctuation between adjacent segments, concentrating compute where attention is most uncertain. During decoding, we introduce \emph{latent KV cache compression}: rather than commit to a token selection at the end of prefilling, we delay compression until a small number of output tokens have been generated, then include those output tokens in the observation window to re-rank cache entries. This re-ranking corrects the prefill to decoding mismatch identified above. The full pipeline requires no fine-tuning and integrates as a drop-in attention replacement.

Our contributions are:

\begin{itemize}[leftmargin=*]
\item
  We identify two regimes of attention heads, Rigid and Dynamic, by segment-wise entropy, and show their assignment is context-dependent and therefore cannot be captured by offline head profiling.
\item
  We propose EntropyInfer, a training-free framework that uses online entropy fluctuation to allocate per-head prefill budgets and re-ranks KV cache entries with output tokens during decoding.
\item
  EntropyInfer achieves up to 2.39$\times$ end-to-end speedup at context lengths beyond 100k tokens with minimal quality drop on LongBench and InfiniteBench, outperforming baselines including SnapKV, AdaKV and CritiPrefill.
\end{itemize}

%% file: section/sec-preliminary-and-related-work.tex
\section{Related Work}\label{sec:related-work}

We use $Q_h, K_h, V_h \in \mathbb{R}^{N \times d}$ to denote the query, key, and value matrices of attention head $h$ on an input of length $N$, and $A_h = \mathrm{softmax}(Q_h K_h^\top / \sqrt{d})$ for its attention weights. The row-wise entropy $H(a_{h,i:}) = -\sum_j a_{h,i,j} \log a_{h,i,j}$ measures how concentrated the $i$-th query's attention is over the keys and is the central quantity our method exploits. Below we review prior efforts to reduce the cost of computing or storing $A_h$.

\subsection{Prefilling Acceleration}\label{sec:rel-prefill}

Sparse attention methods reduce the quadratic cost of computing $A_h$ by selecting only a subset of entries, and they differ in which axis they vary the budget along. Longformer~\citep{beltagyLongformerLongDocumentTransformer} and BigBird~\citep{zaheerBigBirdTransformers} fix the subset geometrically (local windows plus a handful of global tokens) for every input and every head, sacrificing context adaptivity. MInference~\citep{jiangMInference10Accelerating2024} makes the subset head-aware by maintaining a small library of per-head sparsity templates and assigning a template at runtime, but each template is itself static. FlexPrefill~\citep{laiFlexPrefillCONTEXTAWARESPARSE2025} removes the predefined templates and tunes a single sparsity threshold per layer, recovering context adaptivity but flattening the differences between heads inside that layer. CritiPrefill~\citep{lvCritiPrefillSegmentwiseCriticalitybased2024} preserves head-awareness by partitioning the matrix into blocks and selecting the top-$k$ blocks per segment, yet the same $k$ is shared across all heads in the segment, forcing heads with very different attention concentration patterns into the same compute envelope. Each step relaxes one assumption while retaining another, and no existing method varies the budget along all three axes simultaneously: position, head, and context.
\begin{figure*}[ht]
	\centering
	\begin{subfigure}[b]{0.49\textwidth}
		\includegraphics[width=\textwidth]{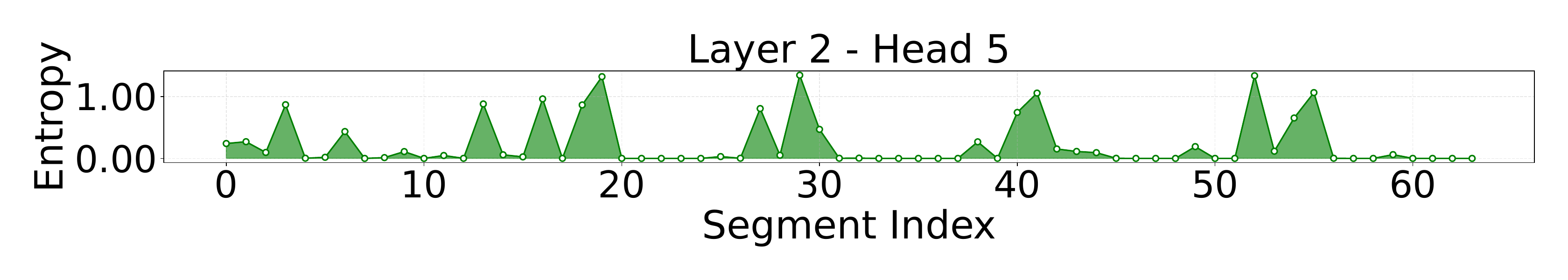}
		\includegraphics[width=\textwidth]{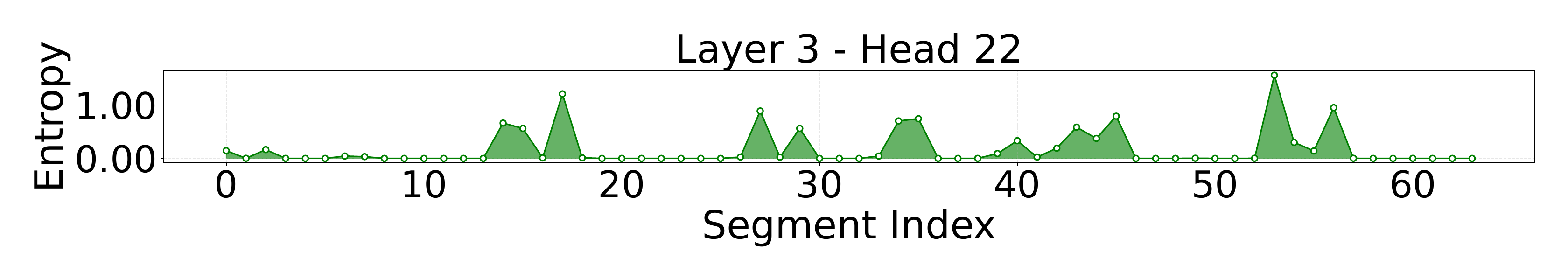}
		\includegraphics[width=\textwidth]{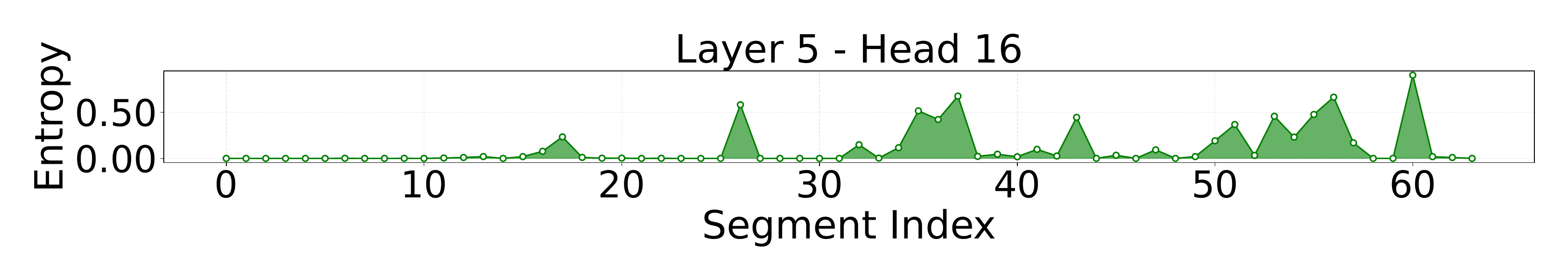}
		\caption{Dynamic heads. Maximum segment entropy $> 10^{-5}$.}
	\end{subfigure}
	\begin{subfigure}[b]{0.49\textwidth}
		\includegraphics[width=\textwidth]{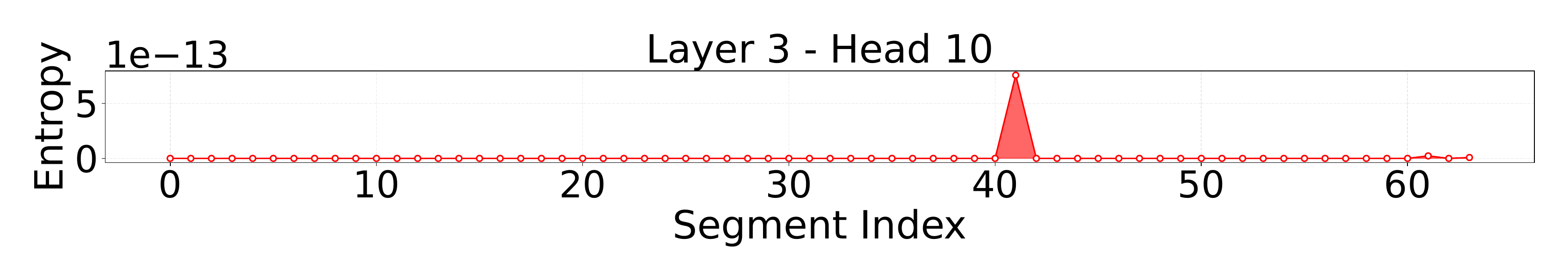}
		\includegraphics[width=\textwidth]{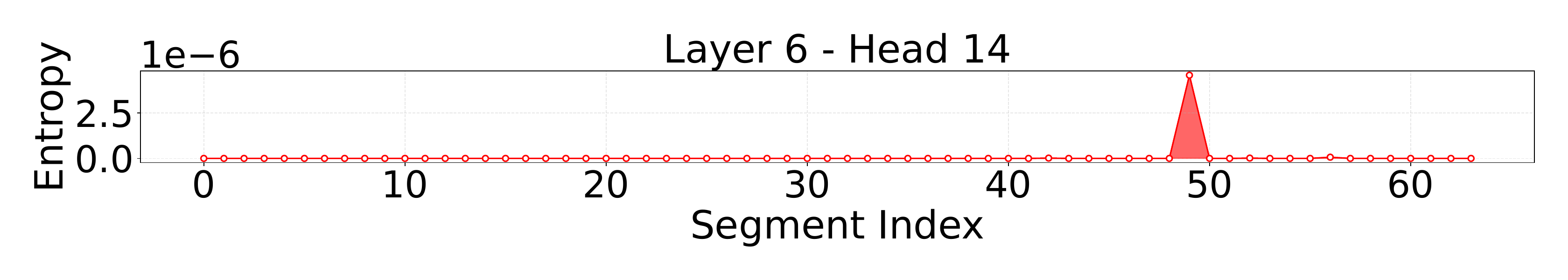}
		\includegraphics[width=\textwidth]{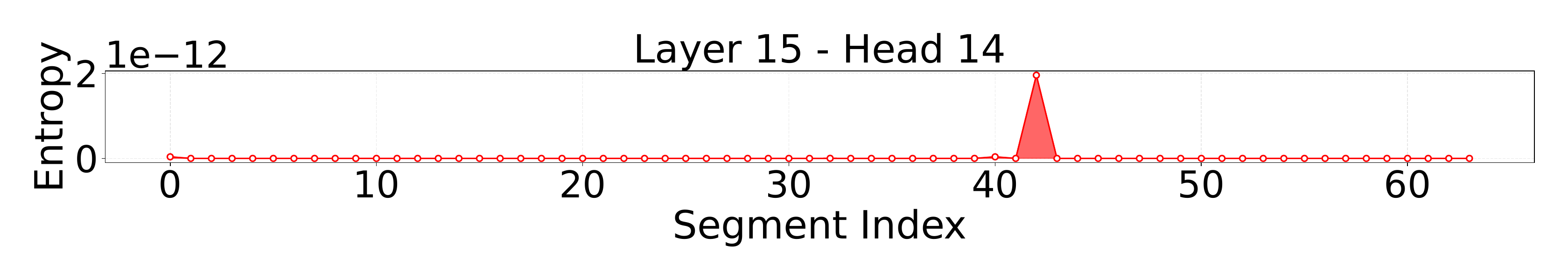}
		\caption{Rigid heads.Maximum segment entropy $\le 10^{-5}$.}
	\end{subfigure}
	\caption{Entropy curves exhibit two distinct patterns across different heads.}
	\label{fig:motivation-entropy-curve}
	\vspace{-1em}
\end{figure*}

\subsection{KV Cache Compression}\label{sec:rel-decode}

Eviction-based methods, the dominant family, compress the KV cache at the end of prefilling by retaining only tokens with the highest attention scores~\citep{Li2025survey}. H2O~\citep{zhangH$_2$OHeavyHitterOracle2023}, SnapKV~\citep{liSnapKVLLMKnows2024}, AdaKV~\citep{fengAdaKVOptimizingKV2024}, and PyramidKV~\citep{yangPyramidInferPyramidKV2024} explore different choices of importance estimator and budget shape: H2O scores tokens against a global heavy-hitter criterion, SnapKV restricts the scoring to an observation window at the end of the input, AdaKV reallocates the per-head budget from that signal, and PyramidKV redistributes the total budget across layers in a pyramidal pattern. StreamingLLM~\citep{xiaoEfficientStreamingLanguage2024} sidesteps the importance question by keeping only sink and recent tokens at the cost of irrecoverable mid-context information.

A second thread tries to improve the importance estimator itself. UNComp~\citep{xiongUNCompCanMatrix2025} and EntropyGuidedKVCaching~\citep{kimEntropyGuidedKVCaching} use the magnitude of attention entropy as a static importance cue at the layer or token level. RazorAttention~\citep{tangRazorAttentionEfficientKV2024} groups attention heads into functional roles (e.g., retrieval versus echo heads) via static analysis prior to inference, and Duo-Attention~\citep{xiaoDuoAttentionEfficientLongContext2024} similarly partitions heads using a learned classifier; both apply role-specific compression rules at runtime. These refinements push the eviction paradigm closer to its quality ceiling but inherit two assumptions from it: prefill-stage attention reliably predicts decoding-stage importance, and head behavior can be characterized before any input is seen. LoopServe and LouisKV~\citep{liLoopServeAdaptiveDualphase2025, wuLouisKVEfficientKV2025} empirically falsify the first by showing that the attention pattern shifts substantially once decoding begins, and our entropy analysis in Section\ref{sec:motivations} falsifies the second by showing that the same head can fall into different regimes on different inputs.

Quantization (KIVI~\citep{liuKIVITuningFreeAsymmetric2023}, KVQuant~\citep{hooperKVQuant10Million2025}) and offloading (FlexGen~\citep{shengFlexGenHighThroughputGenerative2023}, KVSwap~\citep{zhangKVSwapDiskawareKV2025}) reduce KV cost from precision and storage angles orthogonal to selection and can be layered on top of eviction-based methods, including ours. EntropyInfer departs from prior work along both of the assumptions identified above: it categorizes each head online from the entropy signal of the current input, and it defers KV compression past the prefill-decoding boundary so that selection is informed by the tokens the model has begun to generate.

%% file: section/sec-motivation.tex
\begin{figure*}[ht]
	\centering
	\begin{minipage}{0.24\textwidth}
		\centering
		\includegraphics[width=\textwidth]{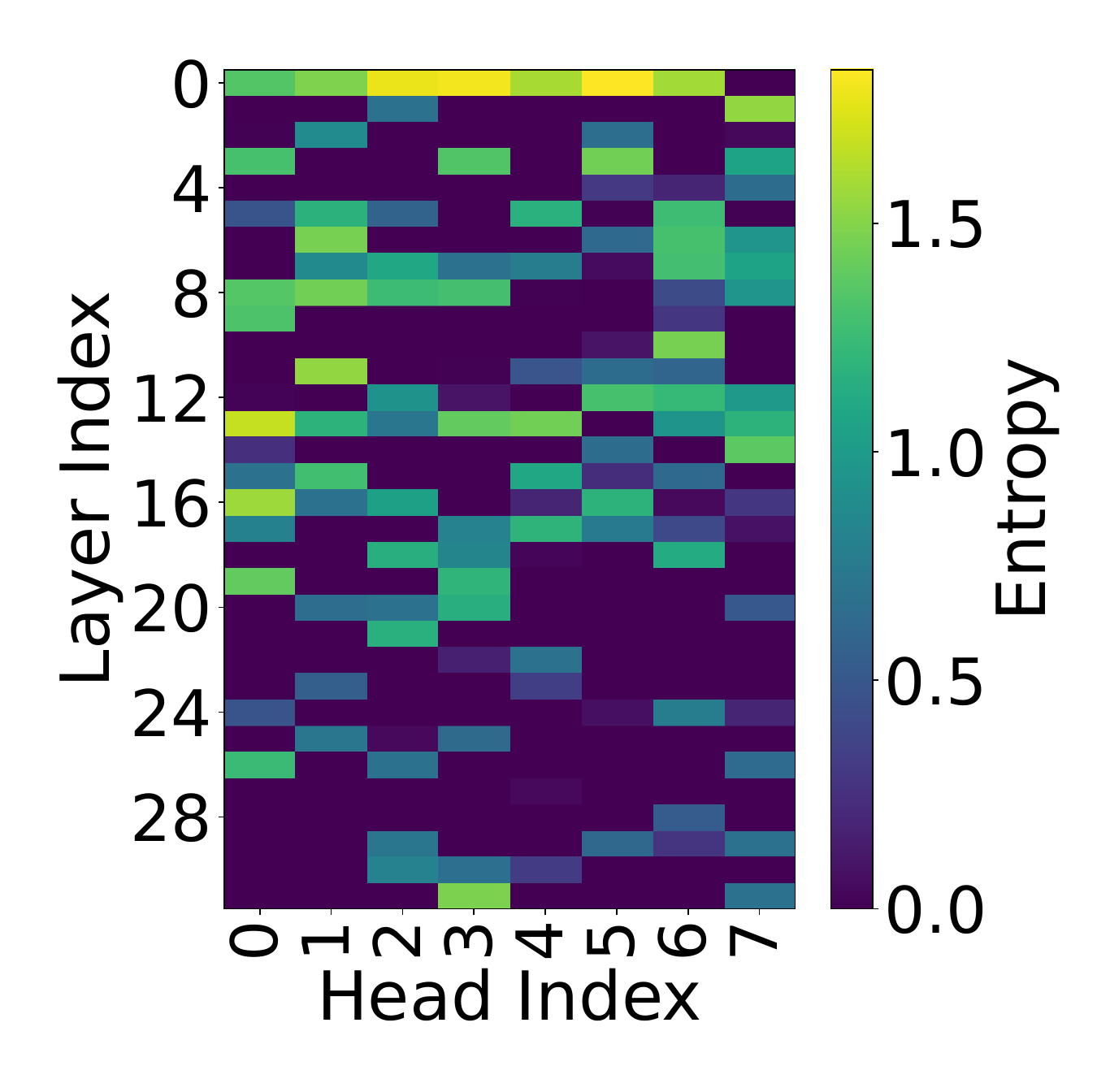}
		\subcaption{GovReport (Averaged).}
	\end{minipage}
	\hfill
	\begin{minipage}{0.24\textwidth}
		\centering
		\includegraphics[width=\textwidth]{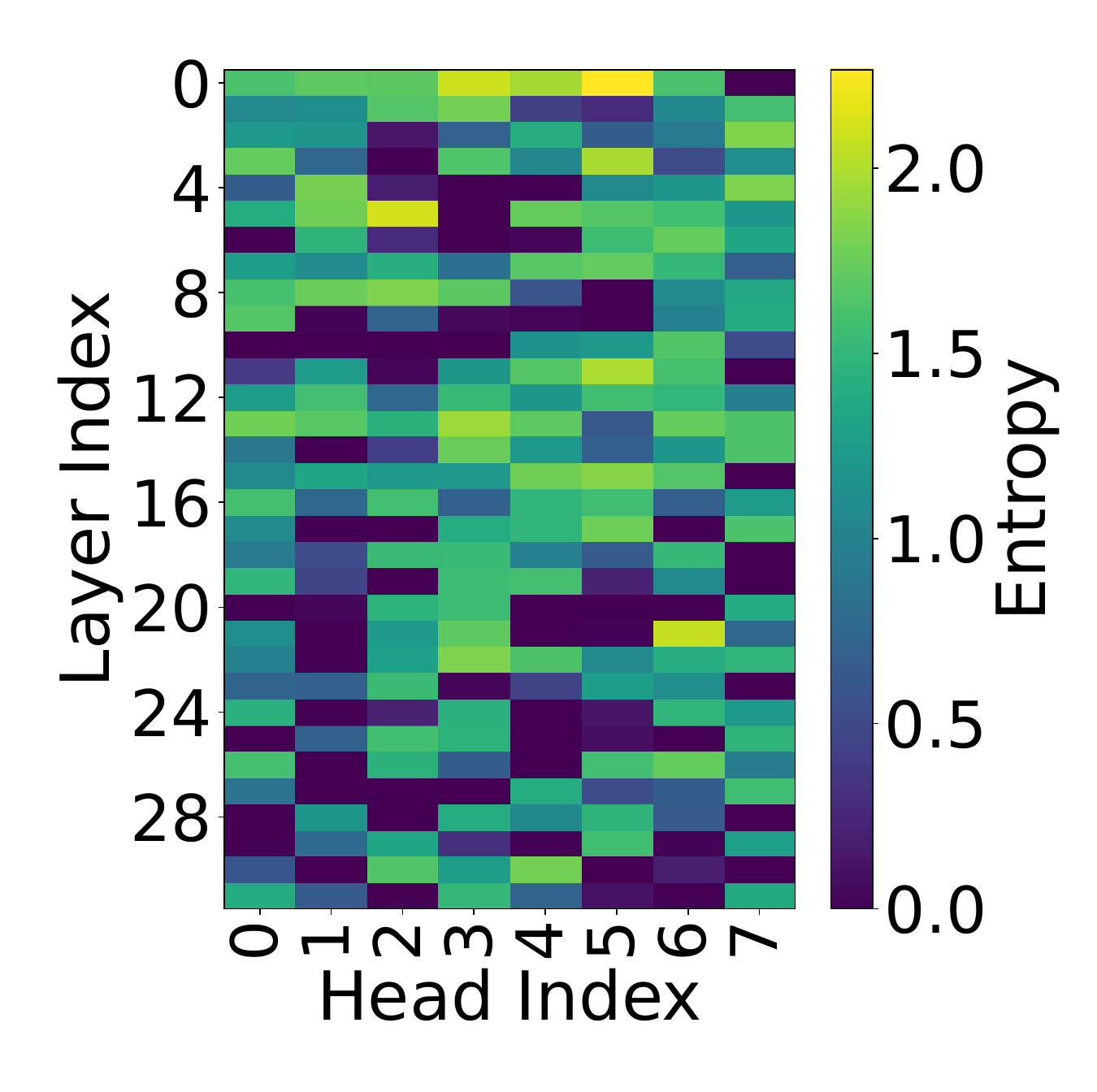}
		\subcaption{Musique (Averaged).}
	\end{minipage}
	\hfill
	\begin{minipage}{0.246\textwidth}
		\centering
		\includegraphics[width=\textwidth]{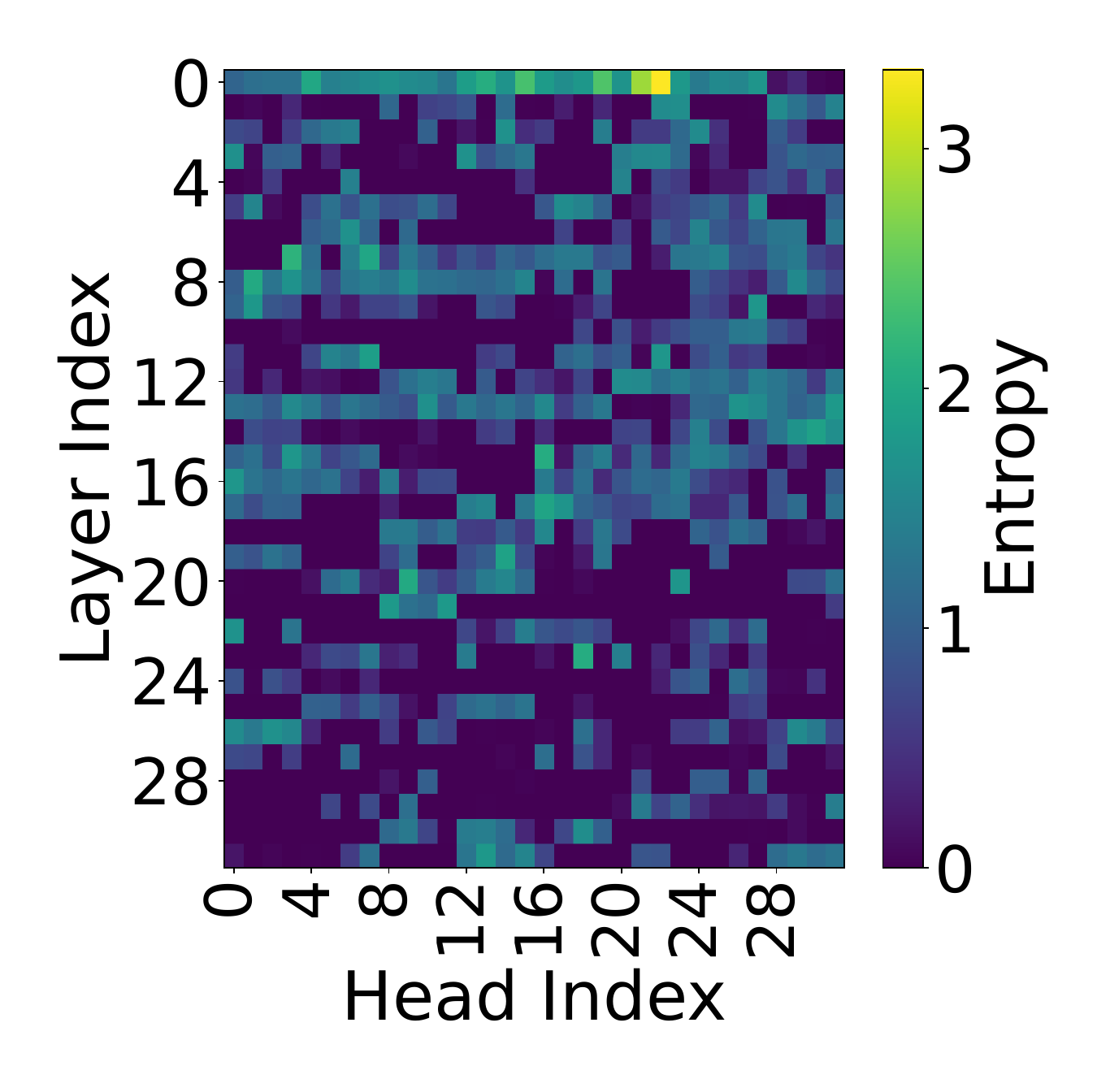}
		\subcaption{GovReport.}
	\end{minipage}
	\hfill
	\begin{minipage}{0.246\textwidth}
		\centering
		\includegraphics[width=\textwidth]{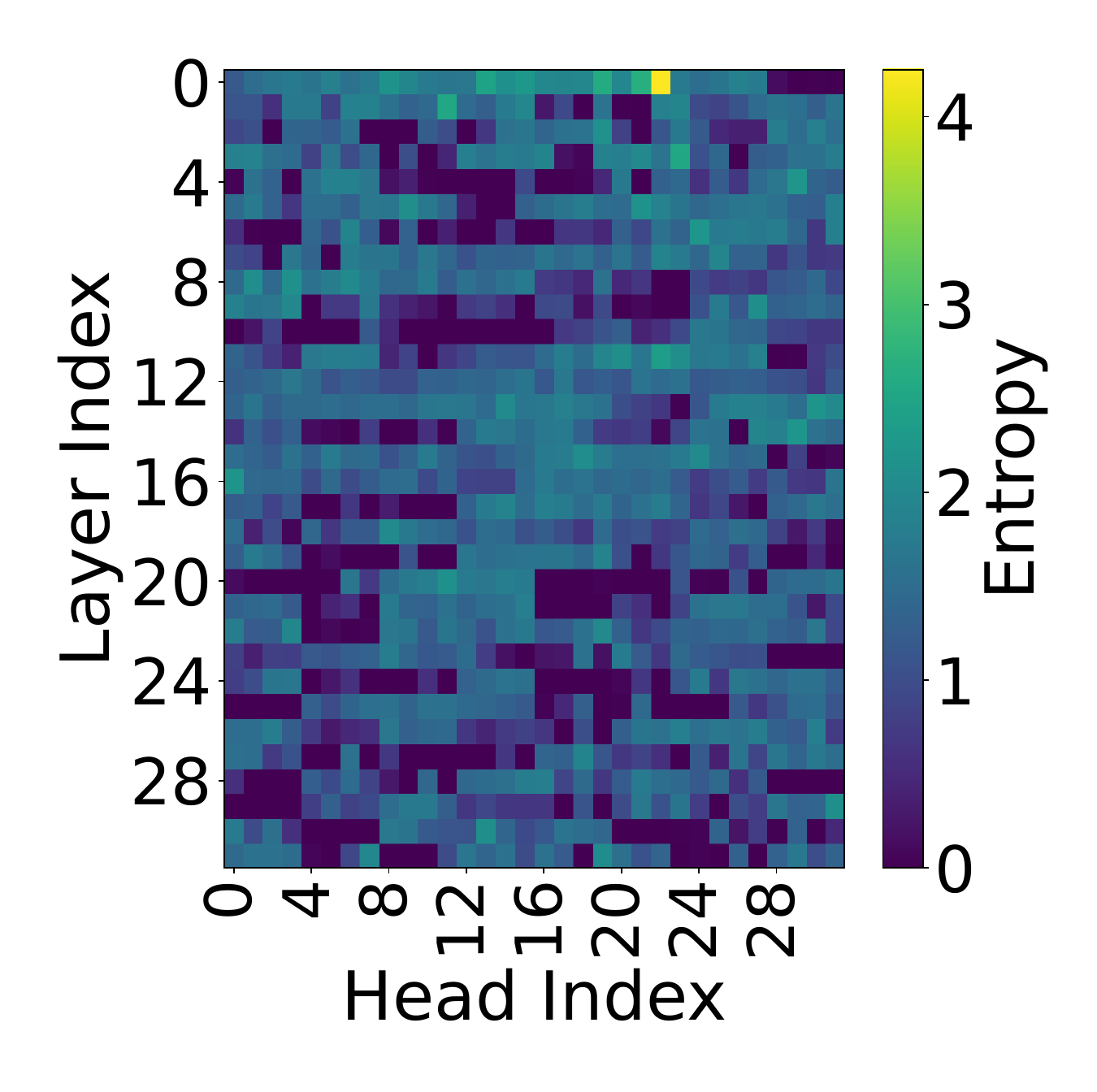}
		\subcaption{Musique.}
	\end{minipage}
	\caption{Entropy pattern varies across heads and different contexts. "Averaged" means aggregating attention entropies per GQA group.}
	\label{fig:entropy-comparison}
	\vspace{-1em}
\end{figure*}

\section{Motivations}\label{sec:motivations}
\subsection{Motivation 1: Attention entropy varies across heads.}\label{motivation-1-attention-entropy-varies-across-heads.}
Previous studies~\citep{kimEntropyGuidedKVCaching, xiongUNCompCanMatrix2025} have revealed the potential for using Shannon's Entropy~\citep{shannonMathematicalTheoryCommunication1948} as a way to measure the degree of attention dispersion. Formally, for query and key sequences \(Q_h  \in \mathcal{R}^{N, d}\) and \(K_h \in \mathcal{R}^{N, d}\) , where \(h\) denotes the index of attention head, \(N\) denotes the sequence length, and \(d\) denotes the hidden dimension, we have the attention weight matrix \(A_h = softmax(\frac{QK^T}{\sqrt{d_k}}) \in \mathcal{R}^{N \times N}\). For each row of \(A_h\) , we can measure the dispersion level of the attention between each query with all keys by calculating the entropy value for each row. Formally, for the \(i_{th}\) row, the entropy is calculated as \(H(a_{i:}) = -\sum^{N}_{j=1} a_{i,j}\log{a_{i,j}}\), where \(a_{i,j}\) corresponds to the attention weight between the \(i_{th}\) query and the \(j_{th}\) key. The higher \(H(a_{i:})\) means the query attends to all keys more evenly, indicating a more evenly distributed importance, while a lower value indicates the concentration of attention on a specific token.

To look into how attention entropy shifts across different queries, we conduct an experiment to plot each head's entropy-query curve. The results are shown in Figure \ref{fig:motivation-entropy-curve}. The curves exhibit two distinct patterns. For the first pattern, the entropy is always below a very small value (e.g.~\(10^{-5}\)), indicating dense and relatively rigid attention distribution. For another pattern, the entropy shifts dynamically with different query, indicating a dynamic pattern of semantic information in this head. We refer the first type of heads as \emph{\textbf{Rigid Heads}}, and the second type of heads as \textbf{\emph{Dynamic Heads}}. Such difference indicates that the responsibilities of different heads in the model are heterogeneous, as also discovered in previous works \citep{tangRazorAttentionEfficientKV2024}.

\subsection{Motivation 2: The sparsity level of heads shifts across different contexts.}\label{motivation-2-the-sparsity-level-of-heads-shifts-across-different-contexts.}

To further understand the impact of different contextual information on the attention entropy distribution across different heads, we visualize the distribution of different type of heads under different contexts. The experiment is conducted on Llama-3.1-8B-Instruct. Following Algorithm \ref{alg:representation-vec} and Algorithm \ref{alg:minmax-attn}'s practice, we segment the input key and query, obtaining an observation attention matrix, and visualize each head's maximum attention entropy across all segments. The results indicate that the distribution of Dynamic Heads and Rigid Heads is context-dependent, highlighting the need for determining the type of the attention head online.

%% file: section/sec-method.tex
\begin{figure*}[ht]
	\centering
	\includegraphics[width=0.9\textwidth]{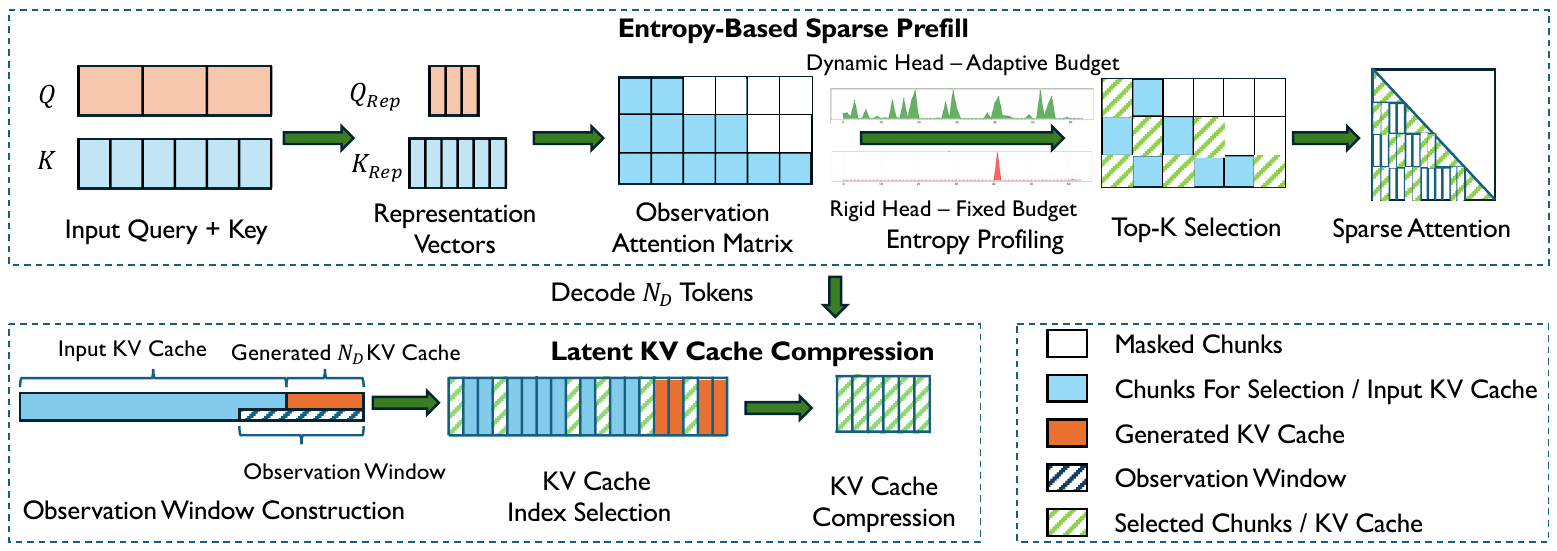}
	\caption{Framework Overview.}
	\label{fig:framework-overview}
	\vspace{-1em}
\end{figure*}

\section{Methods}

In this section, we describe the algorithm used in EntropyInfer. The overall framework is demonstrated in Figure \ref{fig:framework-overview}.

\subsection{Key Notations}

To formalize the framework, we define several key notations. The input context length is denoted by $N$. $L_S$ and $C_S$ refers to segment size and segment count, while $L_B$ refers to block size. Each segment corresponds to different block count and budget, denoted as $C_{B_i}$ and $B_i$ respectively. $B_h$ denotes the base budget for entropy-based sparse prefill and $B_d$ denotes budget for latent KV cache eviction in decode stage.

\subsection{Entropy-Guided Sparse Prefill}\label{entropy-guided-sparse-prefill}

We leverage the entropy information to dynamically determine the importance of different part of the attention matrix. Since calculating attention entropy for each row of the matrix incurs quadratic computation complexity and therefore not realistic, we follow the prior work \citep{lvCritiPrefillSegmentwiseCriticalitybased2024} to break the attention matrix into blocks to create a smaller ``observation attention matrix'', denoted as \(A_{obs}\). The algorithm for creating the matrix is described in Algorithm \ref{alg:representation-vec} and Algorithm \ref{alg:minmax-attn}.

\begin{algorithm}[ht]
	\caption{Calculation of $Q_{Rep}$. $K_{Rep}$ follows similar process.}
	\label{alg:representation-vec}
	\small
	\begin{algorithmic}[1]
		\Require Query States $Q$, Segment Size $L_S$
		\Ensure Representation Vector $Q_{Rep}$
		\State $Q_{Max} \gets Q.\operatorname{reshape}(N_S, L_S, d).\max(\text{dim}=1)$
		\State $Q_{Min} \gets Q.\operatorname{reshape}(N_S, L_S, d).\min(\text{dim}=1)$
		\State $Q_{Rep} \gets \{Q_{Max}, Q_{Min}\}$
		\State \Return $Q_{Rep}$
	\end{algorithmic}
\end{algorithm}

\begin{algorithm}[ht]
	\caption{Observation Attention Calculation}
	\label{alg:minmax-attn}
	\small
	\begin{algorithmic}[1]
		\Require Query States $Q$, Segment Size $L_S$
		\Ensure Observation Attention $Attn_{Obs}$
		\State $Q_{max}, Q_{min} \gets Q_{Rep}$
		\State $K_{max}, K_{min} \gets K_{Rep}$
		\State $S_1 \gets \operatorname{Softmax}(Q_{\max} K_{\max}^T)$
		\State $S_2 \gets \operatorname{Softmax}(Q_{\max} K_{\min}^T)$
		\State $S_3 \gets \operatorname{Softmax}(Q_{\min} K_{\max}^T)$
		\State $S_4 \gets \operatorname{Softmax}(Q_{\min} K_{\min}^T)$
		\State $A^{Obs} \gets \max\left(\frac{S_1 + S_2}{2}, \frac{S_3 + S_4}{2}\right)$
		\State \Return $A^{Obs}$
	\end{algorithmic}
\end{algorithm}

After we have obtained \(A^{obs}\), we calculate each row's attention entropy \(H(a_{i:}) = -\sum^{N}_{j=1} a_{i,j}\log{a_{i,j}}\). As shown in Algorithm \ref{alg:entropy-based-prefill}. To effectively determine the optimal numbers of blocks for each row to reserve, we utilize the fluctuation of the attention entropy across different rows, inspired by prior work \citep{xiongUNCompCanMatrix2025}. We categorize the attention entropy into two categories, the Dynamic Head and the Rigid Head. They're determined by whether the highest attention entropy value exceeds a threshold \(e_t\) . For rigid head, the fluctuation of attention entropy is less informative, making it more suitable to allocate fixed budget for each row. Dynamic head, on the contrary, can be semantically described using the attention entropy information. We set an initial budget for the first row, and for the subsequent rows, we measure the fluctuation level of its attention entropy compared to the previous rows. The higher fluctuation rate means more drastic change in its semantic information interpretation, therefore requiring more budget for precise comprehension. Lower fluctuation rate, on the contrary, reveals a smooth pattern in the semantic information interpretation, suggesting applying similar budget with the previous row for semantic consistency.

\begin{algorithm}[ht]
	\caption{Budget Allocation based on Head Entropy}
	\label{alg:entropy-based-prefill}
	\small
	\begin{algorithmic}[1]
		\Require Head entropy $\{e_i\}$ for $i \in [0, C_S]$, Base budget $B_h$
		\Ensure Allocated budget $\{B_i\}$ for each segment
		\State $e_t=10^{-5}$ \Comment{Entropy threshold}
		\State $\alpha=0.5$ \Comment{Budget coefficient}
		\State $\Delta_t=0.4$ \Comment{Variance percentage threshold}
		\State \emph{\% Categorize heads into 2 types: Rigid head and Dynamic head}
		\If{$\max(\{e_i\}) < e_t$} \Comment{Rigid head (e.g., $e_t = 10^{-5}$)}
		\State $B_i \gets B_h, \quad \forall i \in [0, \text{segment\_count}]$
		\Else \Comment{Dynamic head ($\max(\{e_i\}) \geq e_t$)}
		\State $B_0 \gets B_h$
		\For{$i = 1 \textbf{ to } \text{segment\_count}$}
		\State $\Delta_{\text{budget}} \gets \alpha \cdot B_h \cdot \left( \frac{|e_i - e_{i-1}|}{e_{i-1}} - \Delta_t \right)$
		\State $B_i \gets B_{i-1} + \Delta_{\text{budget}}$
		\State $B_i \gets \max(\min(B_i, B_0), 3\cdot B_0)$
		\EndFor
		\EndIf
		\State \Return $\{B_i\}$
		
	\end{algorithmic}
\end{algorithm}

\subsection{Latent KV cache compression during decoding stage}\label{latent-kv-cache-compression-during-decoding-stage}

To mitigate the growing size of the KV cache, it's common practice to compress KV cache during the generation process. Existing works~\citep{liSnapKVLLMKnows2024, fengAdaKVOptimizingKV2024} compress the KV cache right after the prefilling stage, and utilize an observation window at the end of the input to select important tokens. Such practice is not satisfying, since recent studies have revealed that output tokens is more effective in selecting important tokens during generation, and has distinct attention pattern compared to input tokens~\citep{liLoopServeAdaptiveDualphase2025, wuLouisKVEfficientKV2025}. We design an algorithm to dynamically select important KV cache after generating $N_d$ tokens, as shown in Algorithm \ref{alg:kv_compression}.

\subsection{Complexity Analysis}\label{complexity-analysis}

The attention computation complexity for each attention head is upper-bounded by 
$$
\mathcal{O}((\frac{1}{L_S L_B} + \frac{3}{2L_SL_B} \log \frac{B_h}{L_B}) N^2 + 3B_h N)
$$
Since the coefficient of $N^2$ is negligible compared to dense attention, the attention computation's complexity is nearly linear, as also shown in Figure \ref{fig:e2e-latency-llama}. The proof is detailed in Appendix \ref{appx:proof-complexity}. 

Compared to dense attention computation, which has a complexity of $\mathcal{O}(N^2)$, EntropyInfer exhibits solid efficiency gain, especially in long context scenarios as $N$ grows beyond 100K.

%% file: section/sec-experiment.tex
\section{Experiment}\label{sec-5-experiment}

\begin{table*}[ht]
	\caption{Results on LongBench. CPrefill stands for CritiPretill. Results in bold refers to the best results while underlined results refers to the second best. }
	\label{tab:longbench}
	\setlength{\tabcolsep}{3pt}
	\centering
	\small
	\vspace{-1em}
	\begin{tabular}{c|cccccc|cccccc}
		\hline
		\textbf{Model}            & \multicolumn{6}{c|}{\textbf{Llama-3.1-8B-Instruct}}                                                                   & \multicolumn{6}{c}{\textbf{Qwen2.5-7B-Instruct}}                                                                         \\ \hline
		\textbf{Method}           & \multicolumn{1}{c|}{Base}  & Adakv          & CPrefill      & SnapKV         & \textbf{Ours}  & \textbf{\makecell{Ours \\w/o LD}} & \multicolumn{1}{c|}{Base}   & AdaKV           & CPrefill    & SnapKV          & \textbf{Ours}  & \textbf{\makecell{Ours \\w/o LD}} \\
		\hline
		\textbf{NrtvQA}           & \multicolumn{1}{c|}{29.71} & 29.53          & 27.29          & 29.64          & {\ul 29.86}    & \textbf{30.20}       & \multicolumn{1}{c|}{28.73}  & {\ul 28.43}     & 25.44          & \textbf{28.57}  & 26.81          & 26.81                \\
		\textbf{Qsp}              & \multicolumn{1}{c|}{45.30} & 42.91          & \textbf{45.43} & 43.89          & 45.18          & {\ul 45.38}          & \multicolumn{1}{c|}{44.21}  & 42.20           & \textbf{44.20} & 42.61           & 43.98          & {\ul 44.14}          \\
		\textbf{MFQA\_en}         & \multicolumn{1}{c|}{55.13} & 53.70          & \textbf{55.46} & 54.12          & {\ul 54.74}    & 55.20                & \multicolumn{1}{c|}{53.05}  & {\ul 51.86}     & 51.19          & 51.34           & 51.73          & \textbf{52.13}       \\
		\textbf{HotpotQA}         & \multicolumn{1}{c|}{55.50} & 55.23          & \textbf{56.54} & 54.72          & 55.28          & {\ul 55.37}          & \multicolumn{1}{c|}{57.98}  & 56.95           & 54.65          & 57.16           & {\ul 57.20}    & \textbf{57.30}       \\
		\textbf{2WikiMQA}         & \multicolumn{1}{c|}{44.30} & {\ul 44.84}    & 43.62          & 44.81          & \textbf{45.88} & 44.25                & \multicolumn{1}{c|}{46.33}  & 44.54           & 45.92          & 44.54           & {\ul 47.09}    & \textbf{47.16}       \\
		\textbf{Musique}          & \multicolumn{1}{c|}{31.32} & 30.62          & 29.38          & 29.99          & \textbf{31.74} & {\ul 31.68}          & \multicolumn{1}{c|}{30.03}  & 29.77           & 27.81          & 29.54           & {\ul 30.85}    & \textbf{30.92}       \\
		\textbf{GovReport}        & \multicolumn{1}{c|}{35.16} & 28.36          & {\ul 35.04}    & 28.62          & 30.02          & \textbf{35.18}       & \multicolumn{1}{c|}{31.86}  & 26.60           & {\ul 31.45}    & 27.05           & 28.74          & \textbf{31.48}       \\
		\textbf{QMSum}            & \multicolumn{1}{c|}{25.31} & 24.11          & 24.96          & 24.20          & {\ul 24.97}    & \textbf{25.27}       & \multicolumn{1}{c|}{23.23}  & 23.11           & {\ul 23.19}    & 23.11           & 23.18          & \textbf{23.60}       \\
		\textbf{MultiNews}        & \multicolumn{1}{c|}{27.30} & 26.00          & {\ul 27.26}    & 26.05          & 26.21          & \textbf{27.45}       & \multicolumn{1}{c|}{23.96}  & 22.39           & {\ul 23.76}    & 22.75           & 23.38          & \textbf{23.82}       \\
		\textbf{TREC}             & \multicolumn{1}{c|}{72.50} & 68.50          & 72.00          & 67.50          & \textbf{73.50} & {\ul 73.00}          & \multicolumn{1}{c|}{72.00}  & 68.50           & {\ul 70.00}    & 67.50           & \textbf{71.00} & \textbf{71.00}       \\
		\textbf{TriviaQA}         & \multicolumn{1}{c|}{91.64} & 91.48          & 90.98          & \textbf{91.73} & 91.18          & {\ul 91.49}          & \multicolumn{1}{c|}{88.81}  & {\ul 87.52}     & \textbf{89.09} & 86.69           & \textbf{89.09} & \textbf{89.09}       \\
		\textbf{SamSUM}           & \multicolumn{1}{c|}{43.66} & 42.40          & {\ul 44.18}    & 42.67          & \textbf{44.23} & 43.94                & \multicolumn{1}{c|}{45.64}  & 44.91           & \textbf{46.76} & 45.03           & 46.22          & {\ul 46.23}          \\
		\textbf{PsgCount}         & \multicolumn{1}{c|}{6.96}  & \textbf{7.01}  & 4.64           & {\ul 6.82}     & 6.49           & 5.97                 & \multicolumn{1}{c|}{8.00}   & \textbf{8.00}   & {\ul 7.00}     & \textbf{8.00}   & 6.50           & 6.50                 \\
		\textbf{PsgRetrieval\_en} & \multicolumn{1}{c|}{99.50} & \textbf{99.50} & 97.50          & \textbf{99.50} & {\ul 98.50}    & {\ul 98.50}          & \multicolumn{1}{c|}{100.00} & \textbf{100.00} & 91.00          & \textbf{100.00} & {\ul 91.50}    & {\ul 91.50}          \\
		\textbf{LCC}              & \multicolumn{1}{c|}{63.04} & 62.33          & \textbf{63.24} & 62.13          & 62.77          & {\ul 63.23}          & \multicolumn{1}{c|}{60.07}  & 59.55           & 60.27          & 59.75           & {\ul 60.30}    & \textbf{60.40}       \\
		\textbf{Repobench-P}      & \multicolumn{1}{c|}{56.67} & {\ul 56.35}    & 55.55          & 56.24          & 56.20          & \textbf{56.47}       & \multicolumn{1}{c|}{67.26}  & \textbf{65.75}  & {\ul 65.63}    & 65.19           & 65.26          & 65.53                \\ \hline
		\textbf{Avg}              & \multicolumn{1}{c|}{48.94} & 47.68          & 48.32          & 47.66          & {\ul 48.55}    & \textbf{48.91}       & \multicolumn{1}{c|}{48.82}  & 47.51           & 47.34          & 47.43           & {\ul 47.68}    & \textbf{47.98}       \\ \hline
	\end{tabular}
	\vspace{-1.5em}
\end{table*}

\subsection{Experimental Settings}\label{experimental-settings}

\subsubsection{Dataset and Model}\label{dataset-and-model}

For effectiveness evaluation, we use LongBench \citep{baiLongBenchBilingualMultitask2024} and InfiniteBench \citep{zhang$infty$BenchExtendingLong2024} to evaluate the model's effectiveness. Both datasets contain diverse tasks including \textbf{Question Answering, Summarization, Retrieval and Code Generation, etc.} The two datasets can comprehensively evaluate LLM's performance under diverse scenarios and context lengths, with input sequences of beyond 100K tokens in InfiniteBench.

For efficiency evaluation, we modify Needle-in-a-Haystack's \citep{liNeedleBenchCanLLMs2024} prompt to instruct LLM to generate the summary of the context, and stop the generation process once the LLM has generated 100 tokens.

We use Llama-3.1-8B-Instruct \citep{grattafioriLlama3Herd2024} and Qwen-2.5-7B-Instruct \citep{qwenQwen25TechnicalReport2025} as the backbone model for the experiments.

\begin{table*}[ht]
	\caption{Results on InfiniteBench. R stands for Retrieve, M stands for math, C stands for Code. CPrefill stands for CritiPrefill. Results in bold refers to the best results while underlined results refers to the second best.}
	\label{tab:infinibench}
	\setlength{\tabcolsep}{3pt}
	\centering
	\small
 
	\begin{tabular}{c|cccccc|cccccc}
		\hline
		Model     & \multicolumn{6}{c|}{Llama-3.1-8B-Instruct}                                                                                 & \multicolumn{6}{c}{Qwen2.5-7B-Instruct}                                                                                \\ \hline
		Method    & \multicolumn{1}{c|}{Base}   & SnapKV          & AdaKV           & CPrefill    & \textbf{Ours}   & \textbf{\makecell{Ours \\w/o LD}} & \multicolumn{1}{c|}{Base}   & SnapKV         & AdaKV          & CPrefill      & \textbf{Ours}  & \textbf{\makecell{Ours \\w/o LD}} \\ \hline
		R.PassKey & \multicolumn{1}{c|}{100.00} & \textbf{100.00} & \textbf{100.00} & \textbf{100.00} & \textbf{100.00} & \textbf{100.00}      & \multicolumn{1}{c|}{100.00} & 85.42          & 83.56          & \textbf{99.15} & {\ul 96.44}    & {\ul 96.44}          \\
		R.Number  & \multicolumn{1}{c|}{99.49}  & 85.59           & 82.37           & \textbf{99.15}  & {\ul 95.42}     & {\ul 95.42}          & \multicolumn{1}{c|}{93.22}  & 4.41           & 4.07           & \textbf{89.49} & {\ul 86.78}    & {\ul 86.78}          \\
		En.Dia    & \multicolumn{1}{c|}{21.00}  & 11.50           & 15.00           & 11.00           & {\ul 16.50}     & \textbf{19.00}       & \multicolumn{1}{c|}{16.00}  & 11.00          & {\ul 12.00}    & 10.50          & {\ul 12.00}    & \textbf{15.00}       \\
		En.Sum    & \multicolumn{1}{c|}{26.32}  & 22.35           & 22.00           & {\ul 24.97}     & 22.02           & \textbf{26.48}       & \multicolumn{1}{c|}{21.95}  & {\ul 19.95}    & \textbf{20.62} & 19.84          & 19.75          & 19.40                \\
		En.MC     & \multicolumn{1}{c|}{66.81}  & \textbf{66.81}  & {\ul 66.38}     & 56.77           & 62.01           & 62.01                & \multicolumn{1}{c|}{46.72}  & \textbf{46.72} & {\ul 46.29}    & 37.99          & 44.10          & 44.98                \\
		En.QA     & \multicolumn{1}{c|}{14.44}  & 11.77           & 12.21           & 12.92           & {\ul 14.02}     & \textbf{14.14}       & \multicolumn{1}{c|}{4.81}   & 4.60           & 4.71           & 4.43           & \textbf{5.26}  & {\ul 4.84}           \\
		Zh.QA     & \multicolumn{1}{c|}{13.22}  & 12.03           & 12.05           & 11.03           & {\ul 12.46}     & \textbf{12.70}       & \multicolumn{1}{c|}{9.12}   & \textbf{9.32}  & {\ul 9.18}     & 8.39           & 9.02           & 8.64                 \\
		M.Find    & \multicolumn{1}{c|}{33.14}  & 33.14           & 32.86           & {\ul 34.86}     & \textbf{35.14}  & \textbf{35.14}       & \multicolumn{1}{c|}{38.57}  & 35.14          & {\ul 35.71}    & 34.57          & \textbf{44.57} & \textbf{44.57}       \\
		C.Debug   & \multicolumn{1}{c|}{22.08}  & 22.08           & 22.08           & \textbf{26.65}  & {\ul 25.38}     & {\ul 25.38}          & \multicolumn{1}{c|}{26.14}  & \textbf{25.63} & \textbf{25.63} & {\ul 24.11}    & 23.35          & 23.35                \\ \hline
		Avg       & \multicolumn{1}{c|}{44.06}  & 40.59           & 40.55           & 41.93           & {\ul 42.55}     & \textbf{43.36}       & \multicolumn{1}{c|}{39.62}  & 26.91          & 26.86          & 36.50          & {\ul 37.92}    & \textbf{38.22}       \\ \hline
	\end{tabular}
	\vspace{-1em}
\end{table*}

\subsubsection{Baselines}\label{baselines}

We compare EntropyInfer with the following SOTA methods, covering KV cache compression and sparse prefilling optimizations. 

\begin{itemize}
\tightlist
\item
  \textbf{SnapKV} \citep{liSnapKVLLMKnows2024} is a method that compress KV cache with a snapshot machanism, which reduces inference latency and better utilize memory. 
\item
  \textbf{AdaKV} \citep{fengAdaKVOptimizingKV2024} reallocates budgets across different attention heads, dynamically allocates more cache budget for attention heads with more dispersed patterns.
\item
  \textbf{CritiPrefill} \citep{lvCritiPrefillSegmentwiseCriticalitybased2024} is a method that utilize a criticality-based prefilling method to accelerate the prefill stage.
\end{itemize}

\subsubsection{Evaluation Metrics}\label{evaluation-metrics}

The metrics used by LongBench and InfiniteBench's datasets are detailed as follows.

\begin{itemize}
\item 
For LongBench's metrics, NarrativeQA, Qasper, MultifieldQA-en, HotpotQA, 2WikiMQA and Musique use F1 Score, MultiNews, GovReport and QMSum use Rouge-L Score, TREC, SAMSum, Passage Count and Passage Retrieval use Accuracy, LCC and RepoBench-P use Edit Sim. 
\item 
For InfiniteBench's metrics, En.QA and Zh.QA use F1 Score, En.Sum uses Rouge-L-Sum, En.MC, En.Dia, Code.Debug and Retrieve.PassKey use Accuracy, Retrieve.Number and Math.Find use exact match.
\end{itemize}

The metrics are explained as follows.

\begin{itemize}
\tightlist
\item
 \textbf{F1 Score:} The harmonic mean of precision and recall, providing a balanced evaluation metric for imbalanced datasets.
\item
  \textbf{Accuracy:} The proportion of correctly predicted instances out of the total number of predictions, reflecting the overall correctness of the model.
\item
  \textbf{ROUGE-L:} A metric that measures the Longest Common Subsequence (LCS) between the generated and reference texts to evaluate sequential similarity. It is calculated as: $\text{ROUGE-L} = \frac{\text{LCS}(C, R)}{|R|}$ where $\text{LCS}(C, R)$ represents the length of the longest common subsequence between the candidate text $C$ and the reference text $R$, and $|R|$ denotes the total length of the reference text.
\item 
  \textbf{Edit Sim}: The edit sim ~\cite{svyatkovskiyIntelliCodeComposeCode2020} calculates the Levenstein distance of two text sequences, which is commonly used in evaluation for code generation scenarios.
\item 
  \textbf{Exact Match}: The metric measures the percentage of model responses that exactly matches the ground truth.
\end{itemize}


\subsubsection{Hyperparameters}\label{hyperparameters}

For SnapKV and AdaKV, we set the kv cache budget to 1024 tokens. For streaming LLM, we set the budget to 4096 with 4 sink tokens. For CritiPrefill, we set the prefill budget of each segment to 2048 tokens. For our method, we set the base budget of sparse prefill to 2048 tokens, and the budget for cache eviction to 1024 tokens. 

\subsubsection{Hardware Settings}\label{hardware-settings}

We use a single NVIDIA H100 80G GPU with 192GB CPU memory and 8 CPU Cores to conduct experiments.

\subsection{Main Results}\label{main-results}

\subsubsection{Longbench Results}\label{longbench-results}

We evaluate 16 datasets from LongBench using our method and other baselines, with the backbone model of Llama-3.1-8B-Instruct and Qwen2.5-7B-Instruct. The results are shown in Table \ref{tab:longbench}.  The results show that our method outperforms all other baselines in both Llama and Qwen in terms of average score, and have narrow effectiveness gaps compared to the base model, while in some datasets (e.g. SAMSum and LCC) even outperforming the base model.

\begin{figure}[ht]
	\centering
	\includegraphics[width=0.46\textwidth]{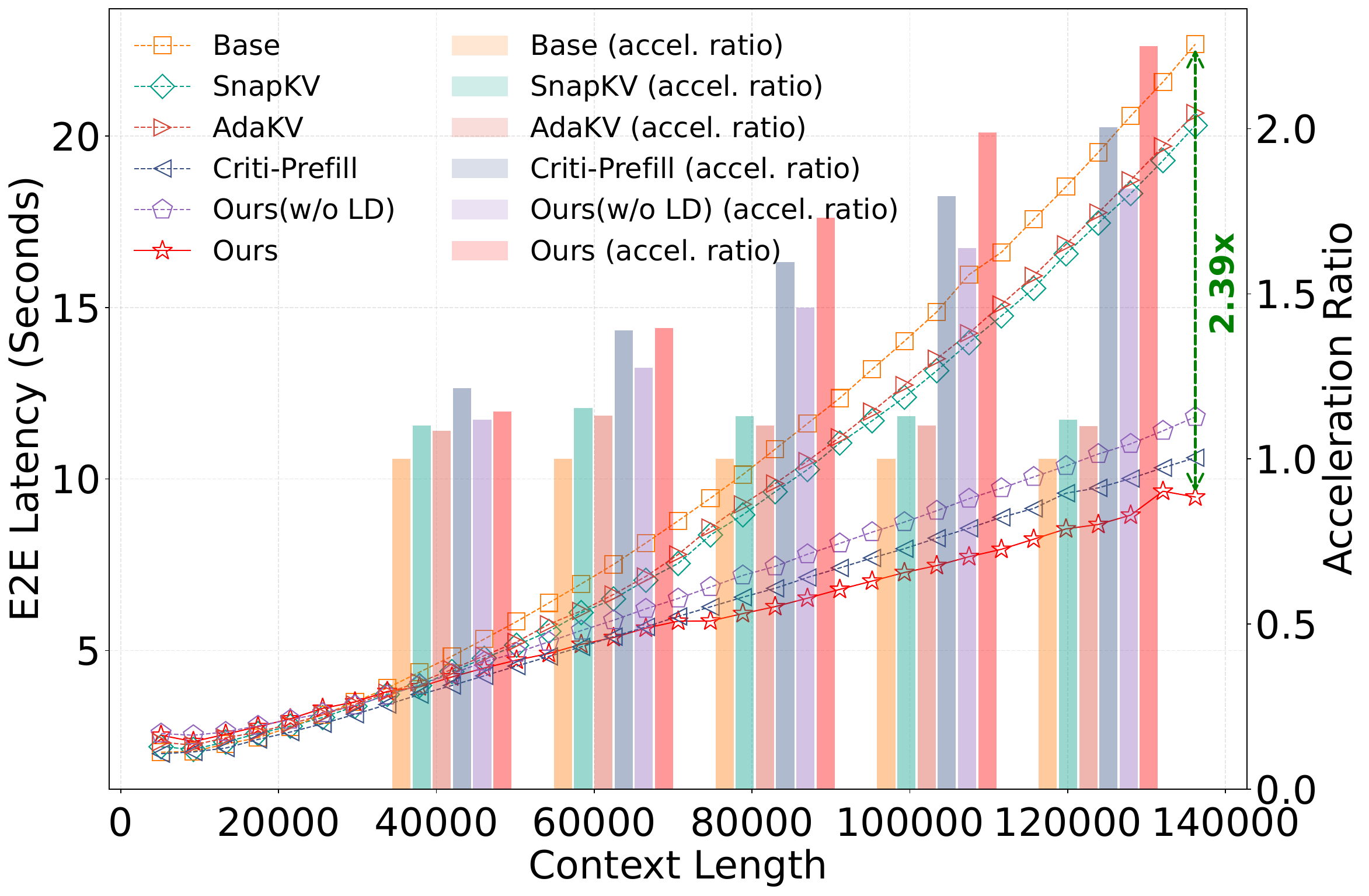}
	\caption{End-to-end latency test under different context lengths on Llama 3.1-8B-Instruct. Lower latency and higher acceleration ratio are better.}
	\label{fig:e2e-latency-llama}
 
\end{figure}

\subsubsection{InfiniteBench Results}\label{infinitebench-results-1}
\begin{figure*}[t]
	\centering
	\begin{minipage}{0.235\textwidth}
		\centering
		\includegraphics[width=\textwidth]{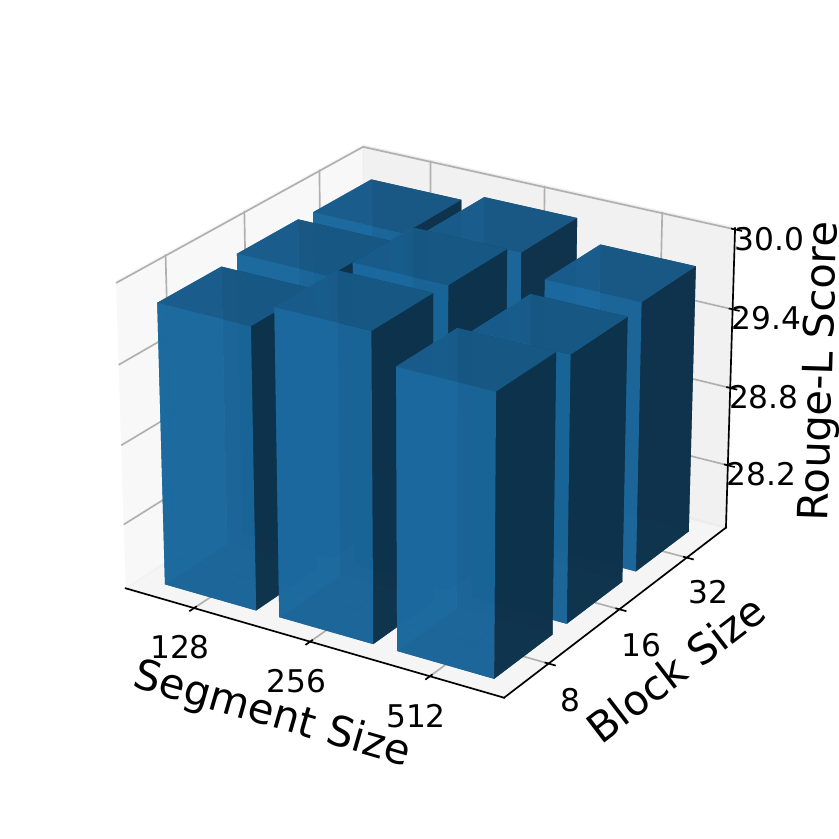}
		\subcaption{GovReport.}
	\end{minipage}
	\begin{minipage}{0.235\textwidth}
		\centering
		\includegraphics[width=\textwidth]{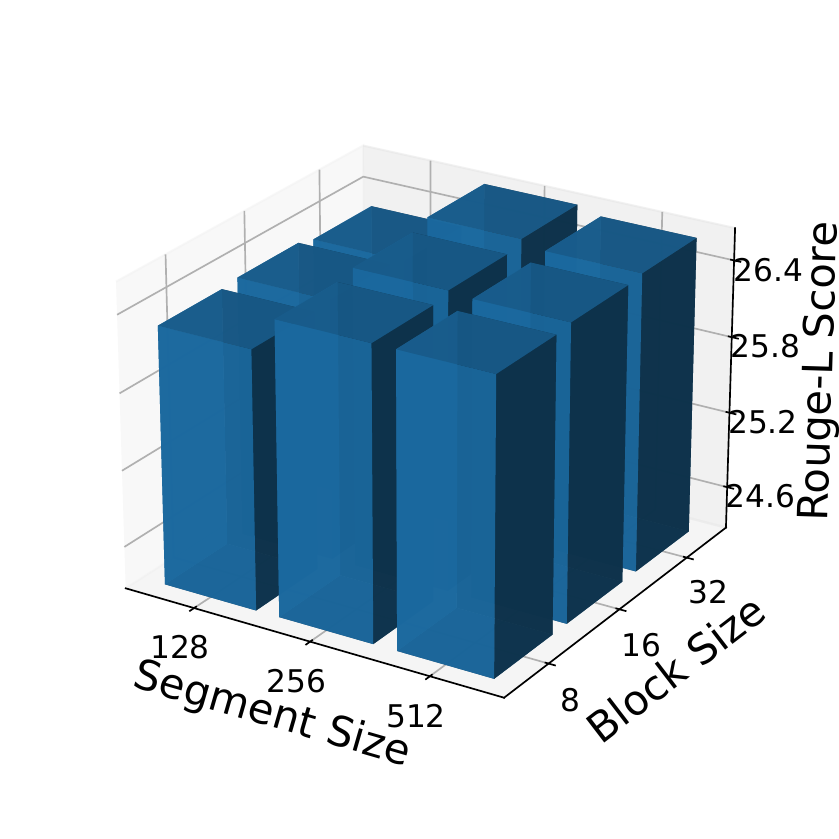}
		\subcaption{MultiNews.}
	\end{minipage}
	\begin{minipage}{0.235\textwidth}
		\centering
		\includegraphics[width=\textwidth]{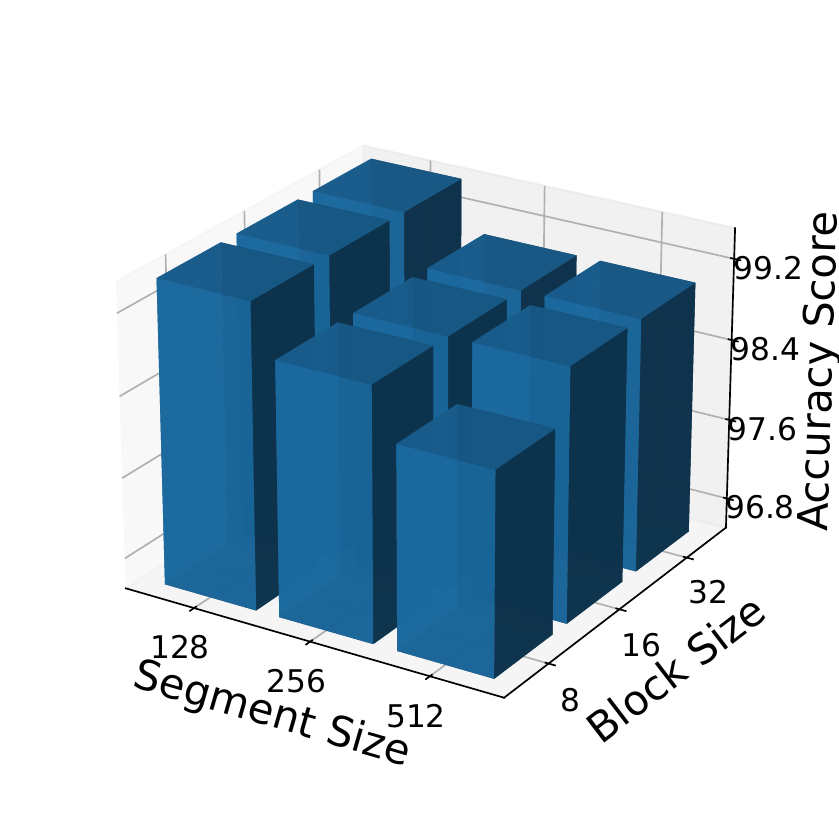}
		\subcaption{Passage-Retrieval.}
	\end{minipage}
	\begin{minipage}{0.235\textwidth}
		\centering
		\includegraphics[width=\textwidth]{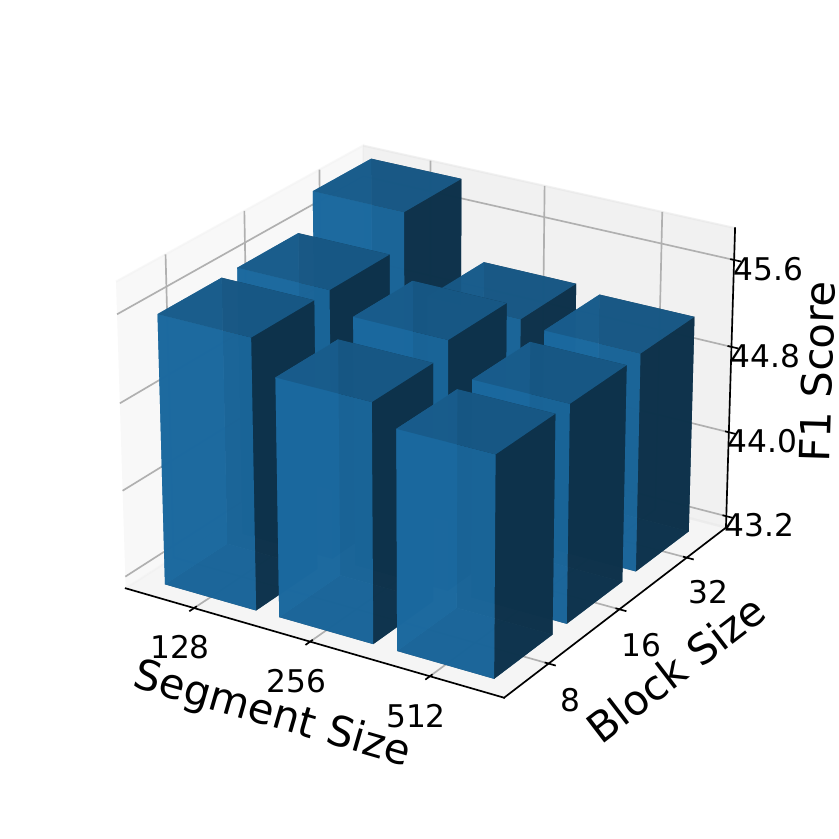}
		\subcaption{Qasper.}
	\end{minipage}
	\caption{Parameter Sensitivity Experiment on Segment and Block Size.}
	\label{fig:param_sensitivity_seg_block_size}
	
\end{figure*}

To evaluate our method's effectiveness on ultra-long context, we conduct experiments on InfiniteBench using Llama-3.1-8B-Instruct~\cite{grattafioriLlama3Herd2024} and Qwen2.5-7B-Instruct~\cite{qwenQwen25TechnicalReport2025}. The results are shown in Table \ref{tab:infinibench}. The results show that our method consistently outperform other baselines on both Llama and Qwen model, highlighting its ability to maintain generation quality even when the input context length is beyond 100K tokens.

\subsection{End-to-end Latency Evaluation}\label{end-to-end-latency}

We evaluate the end-to-end latency using Llama-3.1-8B-Instruct. The model is tasked to summarize essays of different context lengths varying from 4K tokens to 140K tokens, and generate a summary of 100 tokens. Once the generation length reaches 100 tokens, the generation process stops and the end-to-end latency is recorded. 

The results are shown in Figure \ref{fig:e2e-latency-llama}. According to the results, EntropyInfer achieves a solid efficiency gain in long context scenarios compared to other baselines, with a speedup ratio of up to 2.39x. Since EntropyInfer requires performing attention calculation under different budget for each head, its latency is slightly higher than CritiPrefill when using only sparse prefilling (Ours w/o LD), while still greatly outperforming other baselines. With latent decoding enabled, this gap is resolved by reduced decoding latency, making EntropyInfer the most efficient method in long context scenario compared to other baselines.

\subsection{Parameter Sensitivity}\label{parameter-sensitivity}

\subsubsection{Segment and Block Size.}
We perform a grid search to study the parameter sensitivity of Segment Size and Block Size. The results are shown in Figure \ref{fig:param_sensitivity_seg_block_size}. For segment size, larger segment size yields better performance in summary tasks such as GovReport and MultiNews, while retrieval and question answering tasks prefer smaller segment size. For block size, the results remains mostly insensitive, while in some cases larger block size yields better performance. This phenomenon is likely attributed to the consistency of semantic information when using larger block size, as also observed in ~\cite{chenSABlockSemanticAwareKV2025}.

\subsubsection{Prefill and Decode Budget.}

We study the impact of sparse prefill base budget and decode cache budget on EntropyInfer's performance. The results are shown in Figure \ref{fig:ablation-and-param-sensitivity} (b). 

For prefill budget, the results demonstrate that EntropyInfer is essentially insensitive to this hyperparameter in terms of effectiveness. Such insensitivity credits to the adaptive nature of the entropy-based sparse prefill, which dynamically adjusts budget to balance efficiency and generation quality, demonstrating the robustness of the method.
For decode cache budget, the results suggest that higher budgets can lead to better effectiveness performance. However, such effectiveness gain is limited in particular datasets, and would incur extra computation and memory overhead.

\subsection{Ablation Study}\label{ablation-study}

We conduct ablation study using Llama-3.1-8B-Instruct and Longbench. The ablation study involves the following three settings.
\begin{itemize}[leftmargin=*]
	\item \textbf{Ours:} Enable entropy-based sparse prefill and latent decode.
	\item \textbf{Ours w/o SP:} Disable entropy-based sparse attention, enable latent decode.
	\item \textbf{Ours w/o LD:} Enable entropy-based sparse attention, disable latent decode.
\end{itemize}

\begin{figure}[ht]
	\centering
	\begin{minipage}{0.22\textwidth}
		\centering
		\includegraphics[width=\textwidth]{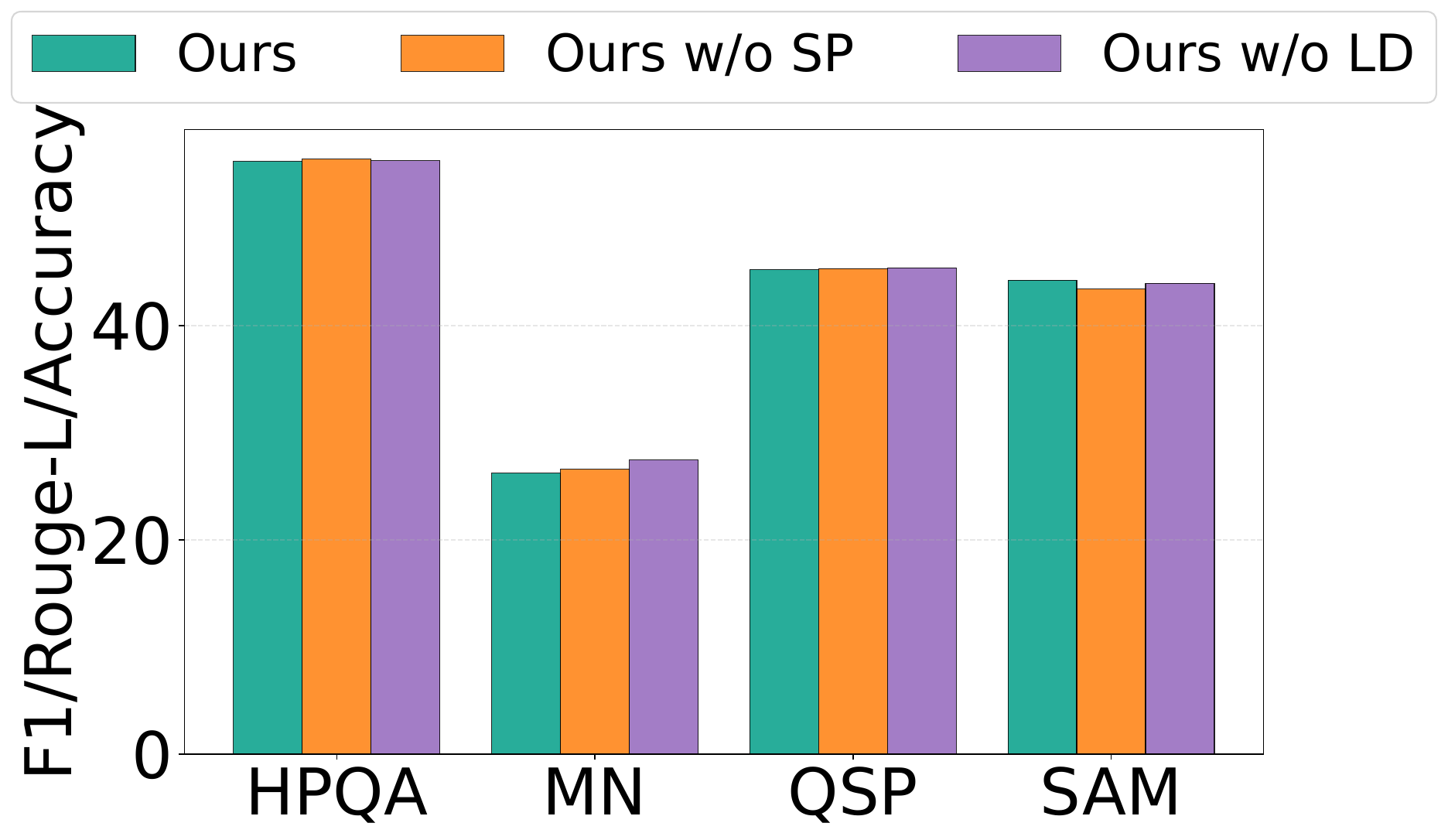}
		\subcaption{Ablation Study.}
	\end{minipage}
	\hfill
	\begin{minipage}{0.25\textwidth}
		\centering
		\includegraphics[width=\textwidth]{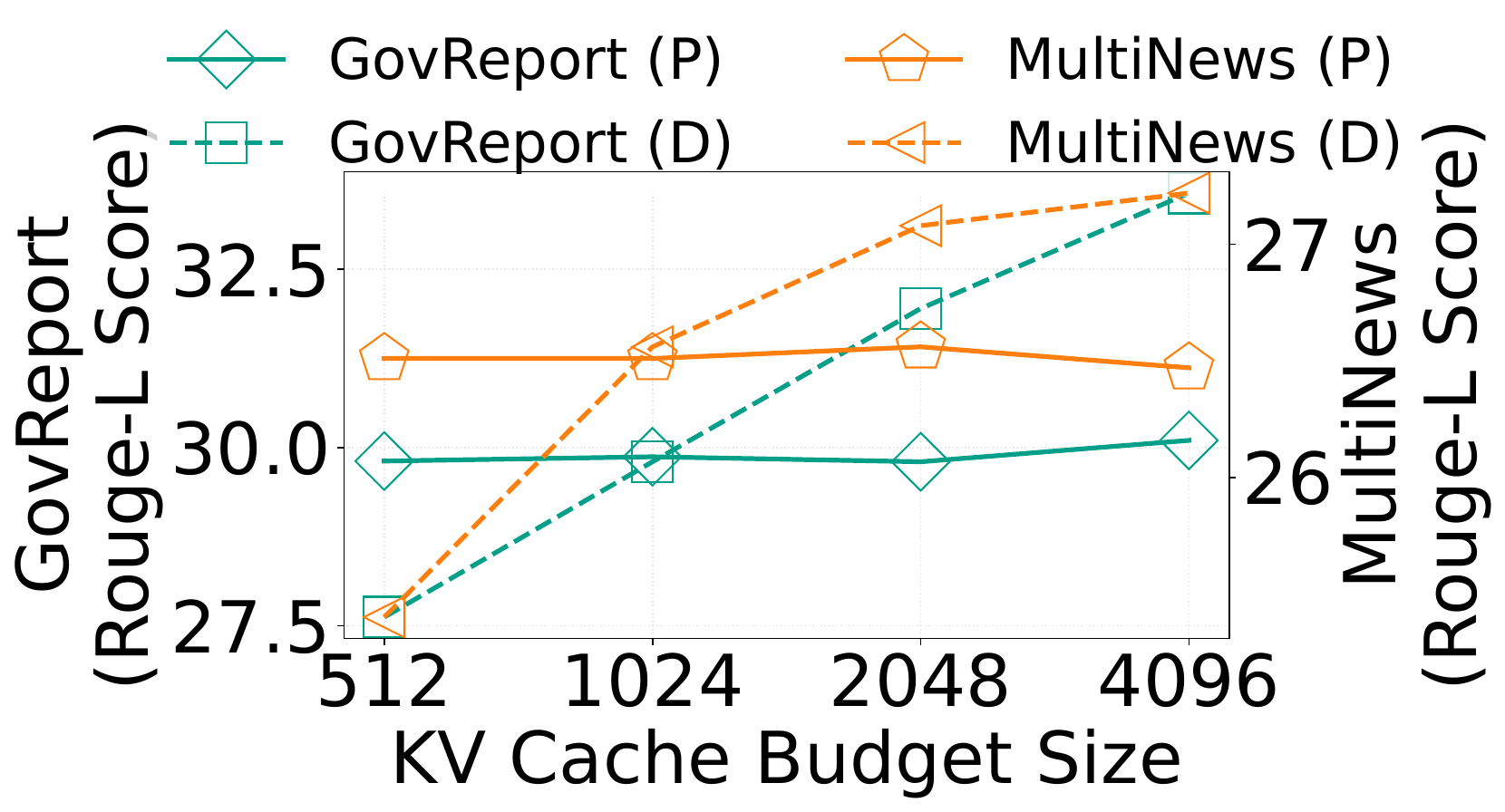}
		\subcaption{Budget Sensitvity.}
	\end{minipage}
	\caption{Ablation Study and Parameter Sensitivity Results. P refers to Prefill, D refers to Decode.}
	\label{fig:ablation-and-param-sensitivity}
 
\end{figure}

The effectiveness results are shown in Figure \ref{fig:ablation-and-param-sensitivity} (a). The result exhibits that enabling latent decode and  entropy-based sparse prefill only have a very slight impact on generation quality, while both modules work in tandem with each other to achieve optimal efficiency results (as shown in Figure \ref{fig:e2e-latency-llama}).

\begin{table*}[ht]
	\centering
	\caption{LongBench results on openPangu-Embedded-1B and openPangu-Embedded-7B-v1.1.}
	\label{tab:longbench-pangu}
	\setlength{\tabcolsep}{1pt}
	\small
	\begin{tabular}{c|c|cccccc|c}
		\hline
		\textbf{Model}                                & \textbf{Method}      & \textbf{NrtvQA} & \textbf{Qasper} & \textbf{MFQA-en} & \textbf{HotpotQA} & \textbf{2WikiMQA} & \textbf{Musique} & \textbf{Avg}   \\ \hline
		\multirow{3}{*}{\textbf{opanPangu-7B}}        & \textbf{Base}        & 14.82           & 37.46           & 44.37            & 35.64             & 29.58             & 20.08            & 30.33          \\
		& \textbf{Ours}        & \textbf{15.12}  & 35.53           & 43.28            & 34.91             & \textbf{31.06}    & 18.04            & 29.66          \\
		& \textbf{Ours w/o LD} & 14.86           & \textbf{36.94}  & \textbf{44.44}   & \textbf{35.23}    & 31.00             & \textbf{18.41}   & \textbf{30.15} \\ \hline
		\multirow{3}{*}{\textbf{openPangu-7B w/ CoT}} & \textbf{Base}        & 15.07           & 37.75           & 37.21            & 47.79             & 61.24             & 31.31            & 38.40          \\
		& \textbf{Ours}        & \textbf{16.10}  & 37.39           & \textbf{36.84}   & 50.84             & 53.91             & 31.94            & 37.84          \\
		& \textbf{Ours w/o LD} & 13.83           & \textbf{38.01}  & 36.43            & \textbf{51.53}    & \textbf{61.29}    & \textbf{32.26}   & \textbf{38.89} \\ \hline
		\multirow{3}{*}{\textbf{openPangu-1B}}        & \textbf{Base}        & 10.36           & 28.75           & 43.66            & 34.69             & 36.05             & 19.63            & 28.86          \\
		& \textbf{Ours}        & 10.45           & 28.86           & 42.75            & 32.90             & \textbf{35.17}    & 18.11            & 28.04          \\
		& \textbf{Ours w/o LD} & \textbf{10.46}  & \textbf{29.38}  & \textbf{43.95}   & \textbf{32.92}    & 35.11             & \textbf{18.17}   & \textbf{28.33} \\ \hline
	\end{tabular}
\end{table*}

\begin{table*}[ht]
	\centering
	\caption{LoCoMo results on openPangu-Embedded-1B and openPangu-Embedded-7B.}
	\label{tab:locomo-pangu}
	\setlength{\tabcolsep}{1pt}
	\small
	\begin{tabular}{c|c|ccccc|c}
		\hline
		\textbf{Model}                                & \textbf{Method}      & \textbf{Multi-Hop} & \textbf{Temporal} & \textbf{Open-domain} & \textbf{Single-Hop} & \textbf{Adversarial} & \textbf{Overall} \\ \hline
		\multirow{3}{*}{\textbf{openPangu-7B}}        & \textbf{Base}        & 27.84              & 16.04             & 10.73                & 31.53               & 3.14                 & 21.12            \\
		& \textbf{Ours}        & \textbf{27.41}     & \textbf{14.66}    & 10.55                & \textbf{31.63}      & 2.91                 & \textbf{20.82}   \\
		& \textbf{Ours w/o LD} & 26.40              & 14.14             & \textbf{10.70}       & 31.51               & \textbf{3.36}        & 20.65            \\ \hline
		\multirow{3}{*}{\textbf{openPangu-7B w/ CoT}} & \textbf{Base}        & 27.16              & 19.07             & 15.52                & 27.50               & 29.82                & 26.03            \\
		& \textbf{Ours}        & \textbf{25.61}     & \textbf{22.58}    & 11.21                & \textbf{27.30}      & 21.08                & 24.12            \\
		& \textbf{Ours w/o LD} & 24.14              & 19.93             & \textbf{13.64}       & 25.10               & \textbf{33.86}       & \textbf{25.54}   \\ \hline
		\multirow{3}{*}{\textbf{openPangu-1B}}        & \textbf{Base}        & 20.08              & 14.16             & 11.69                & 22.32               & 9.64                 & 17.32            \\
		& \textbf{Ours}        & 18.88              & 13.45             & 10.77                & 21.99               & 7.40                 & 16.35            \\
		& \textbf{Ours w/o LD} & \textbf{19.16}     & \textbf{13.53}    & \textbf{10.95}       & \textbf{22.12}      & \textbf{7.85}        & \textbf{16.56}   \\ \hline
	\end{tabular}
\end{table*}

\subsection{Extra expriment on Pangu Model Series.}

\begin{figure}[htbp]
	\centering
	\includegraphics[width=0.4\textwidth]{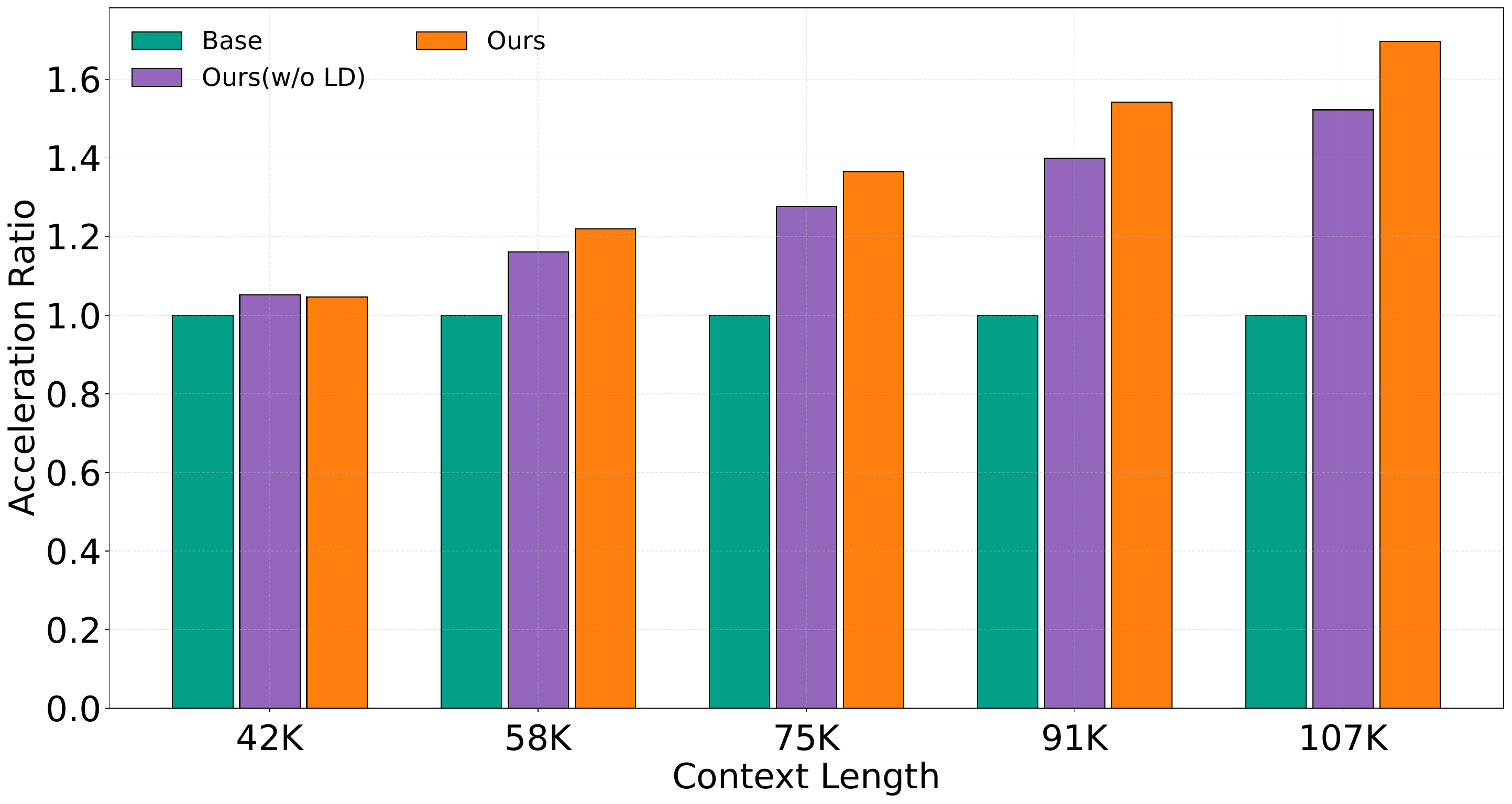}
	\caption{Efficiency results on openPangu-Embedded-7B-v1.1. }
	\label{fig:pangu-efficiency}
\end{figure}

To extensively evaluate our method's adaptibility, we integrate our method into openPangu-Embedded-1B-v1.1 and openPnagu-Embedded-7B-v1.1 ~\cite{chenPanguEmbeddedEfficient2025}. We conduct end-to-end efficiency experiment under different context lengths, as shown in Figure ~\ref{fig:pangu-efficiency}. The results show that our method consistently accelerates model inference under different long context lengths, and the acceletation is more significant for longer contexts. For effectiveness, we use LongBench ~\cite{baiLongBenchBilingualMultitask2024} and LoCoMo~\cite{maharana2024evaluating} to evaluate our method's effectiveness on long context and multi-turn conversation scenarios. The results are shown in Table ~\ref{tab:longbench-pangu} and Table ~\ref{tab:locomo-pangu}. The efficiency and effectiveness results demonstrates that our method achieves a sigificant acceleration in end-to-end generation while maintaining high generation quality.

%% file: section/sec-conclusion.tex
\section{Conclusion}

In this work, we propose EntropyInfer, a method for accelerating LLM inference while preserving generation quality. By utilizing the attention entropy information during prefill stage to dynamically select important attention blocks, and perform latent kv cache compression in decoding stage, EntropyInfer achieves a speedup of up to 2.39x in end-to-end generation while maintaining generation quality. EntropyInfer can serve as an effective solution for long context LLM inference without the need for fine-tuning the models, making it a practical method for real-world applications.

%% file: section/sec-limitation.tex
\section*{Limitations}

We observe limited performance boost in short context inference, likely due to the introduced observation attention calculation and entropy profiling overhead. However, the introduced overhead does not incur significant latency in short context lengths, and the performance gain of sparse prefill and latent decode outweighs the extra overhead as the context length increases. 

\section*{Ethics Statement}
This work does have any ethical issues.

%% file: section/sec-appendix.tex
\appendix

\section{Proof for Complexity Analysis}\label{appx:proof-complexity}

We denote the computation overhead for an input sequence of length $N$ as $T(N)$.

The complexity of the sparse attention process mainly consists of three parts, including observation attention calculation process $T_{obs}(N)$,  Top-K block selection process $T_{topk}(N)$ and sparse attention computation process $T_{spa}(N)$. They're expressed in Equations (\ref{eq:t_obs}) (\ref{eq:t_topk}) and (\ref{eq:t_spa}) separately.

\begin{equation}
	\label{eq:t_obs}
	\begin{aligned}
		&T_{obs}(N) = C_S \times C_B= \frac{N}{L_S} \times \frac{N}{L_B} = \frac{N^2}{L_SL_B}
	\end{aligned}
\end{equation}

\begin{equation}
	\label{eq:t_topk}
	\begin{aligned}
		&T_{topk}(N) = \sum^{C_S}_{i=0}C_{B_i}\log\frac{B_i}{L_B} = \frac{3}{2}\sum^{C_S}_{i=0}C_B\log\frac{B_h}{L_B} = \frac{3}{2}C_SC_B\log\frac{B_h}{L_B}= \frac{3N^2}{2L_SL_B}\log\frac{B_h}{L_B}
	\end{aligned}
\end{equation}

\begin{equation}
	\label{eq:t_spa}
	\begin{aligned}
		&T_{spa}(N) = \sum^{C_S}_{i=0}B_i \times L_S \le 3\sum^{C_S}_{i=0}B_hL_S = 3C_SB_hL_S = 3B_hN
	\end{aligned}
\end{equation}

Therefore, the total complexity $T_{total}(N)$ can be expressed as Equation (\ref{eq:total-complexity}).

\begin{equation}
	\label{eq:total-complexity}
	\begin{aligned}
		&T_{total}(N)=T_{obs}(N) + T_{topk}(N) + T_{spa}(N) = \mathcal{O}((\frac{1}{L_S L_B} + \frac{3}{2L_SL_B} \log \frac{B_h}{L_B}) N^2 + 3B_h N)
	\end{aligned}
\end{equation}

\section{Latent KV Cache Compression Algorithm}

The latent KV cache compression algorithm is detailed in Algorithm \ref{alg:kv_compression}.

\begin{algorithm}[ht]
	\caption{Latent Decode KV Cache Compression via Importance Scoring}
	\label{alg:kv_compression}
	\begin{algorithmic}[1]
		\Require 
		\Statex Original Key states matrix: $K \in \mathbb{R}^{L \times d}$
		\Statex Original Value states matrix: $V \in \mathbb{R}^{L \times d}$
		\Statex Query states matrix: $Q \in \mathbb{R}^{L \times d}$
		\Statex Observation window size: $N_W$, Number of newly inferred tokens: $N_D$
		\Statex Cache budget: $B_d$
		\Ensure 
		\Statex Compressed KV Cache: $\text{KV}_{compressed}$
		\State \textbf{Step 1: Construct observation window index set}
		\State {Determine window indices $\mathcal{I}_W \subset \{1, \dots, L\}$ such that $|\mathcal{I}_W| = N_W$}.
		\State Extract observation window Query matrix: $Q_W \leftarrow Q[\mathcal{I}_W, :]$ \Comment{$Q_W \in \mathbb{R}^{N_W \times d}$}
		\State \textbf{Step 2: Compute global importance scores}
		\State Compute interaction matrix: $S \leftarrow Q_W K^\top$ \Comment{$S \in \mathbb{R}^{N_W \times L}$}
		\State Aggregate to obtain importance vector: $\mathbf{w} \leftarrow \sum_{i=1}^{N_W} S_{i,:}$ \Comment{$\mathbf{w} \in \mathbb{R}^{1 \times L}$}
		\State \textbf{Step 3: Sparse index selection}
		\State Select the indices of the $B_d$ tokens with the highest weights:
		\State $\mathcal{I}_{idx} \gets TopK(\mathbf{w}, B_d)$ 
		\State \Comment{$\mathcal{I}_{idx} = \{j \mid w_j \text{ is among the top } B \text{ largest elements in } \mathbf{w}\}$}
		\State \textbf{Step 4: Cache eviction and compression}
		\State $\text{KV}_{compressed} \leftarrow \{ (K_{j,:}, V_{j,:}) \mid j \in \mathcal{I}_{idx} \}$
		\State \Return $\text{KV}_{compressed}$
	\end{algorithmic}
\end{algorithm}